          [2001/04/25 1.1 (PWD)]
\documentclass[a4paper,twocolumn]{esapub2005} 
\usepackage{url}
\usepackage{graphicx}
\usepackage{amsmath}
\usepackage{booktabs}
\usepackage{array}
\usepackage{tabularx}
\usepackage[table]{xcolor}
\definecolor{headergray}{gray}{0.8}
\definecolor{subheadergray}{gray}{0.88}
\definecolor{rowgray}{gray}{0.95}

\usepackage[
    backend=biber,
    style=ieee,
    natbib=true,
    url=true,
    doi=true,
    eprint=false,
    maxnames=4,
    maxcitenames=2,
    mincitenames=1,
    autocite=superscript
]{biblatex}
\title{THE DISTANT DESIGN FOR REMOTE TRANSMISSION AND STEERING SYSTEMS FOR PLANETARY ROBOTICS}

\author{Cristina Luna}
\author{Alba Guerra}
\author{Almudena Moreno}
\author{Manuel Esquer}
\author{Willy Roa}
\affil{GMV Aerospace and Defence SAU, Spain, cluna@gmv.com}
\author{Mateusz Krawczak}
\affil{PIAP Space, Poland}
\author{Robert Popela}
\affil{Brno University of Technology, Czechia}
\author{Piotr Osica}
\affil{Spacive, Poland}
\author{Davide Nicolis}
\affil{ESA ESTEC, Netherlands}

\begin{document}


\maketitle

\begin{abstract}
Planetary exploration missions require robust locomotion systems capable of operating in extreme environments over extended periods. This paper presents the DISTANT (Distant Transmission and Steering Systems) design, a novel approach for relocating rover traction and steering actuators from wheel-mounted positions to a thermally protected warm box within the rover body. The design addresses critical challenges in long-distance traversal missions by protecting sensitive components from thermal cycling, dust contamination, and mechanical wear. A double wishbone suspension configuration with cardan joints and capstan drive steering has been selected as the optimal architecture following comprehensive trade-off analysis. The system enables independent wheel traction, steering control, and suspension management whilst maintaining all motorisation within the protected environment. The design meets a 50 km traverse requirement without performance degradation, with integrated dust protection mechanisms and thermal management solutions. Testing and validation activities are planned for Q1 2026 following breadboard manufacturing at 1:3 scale.
\end{abstract}

\section{INTRODUCTION}

\begin{figure}[ht!]
    \centering
    \includegraphics[width=1\linewidth]{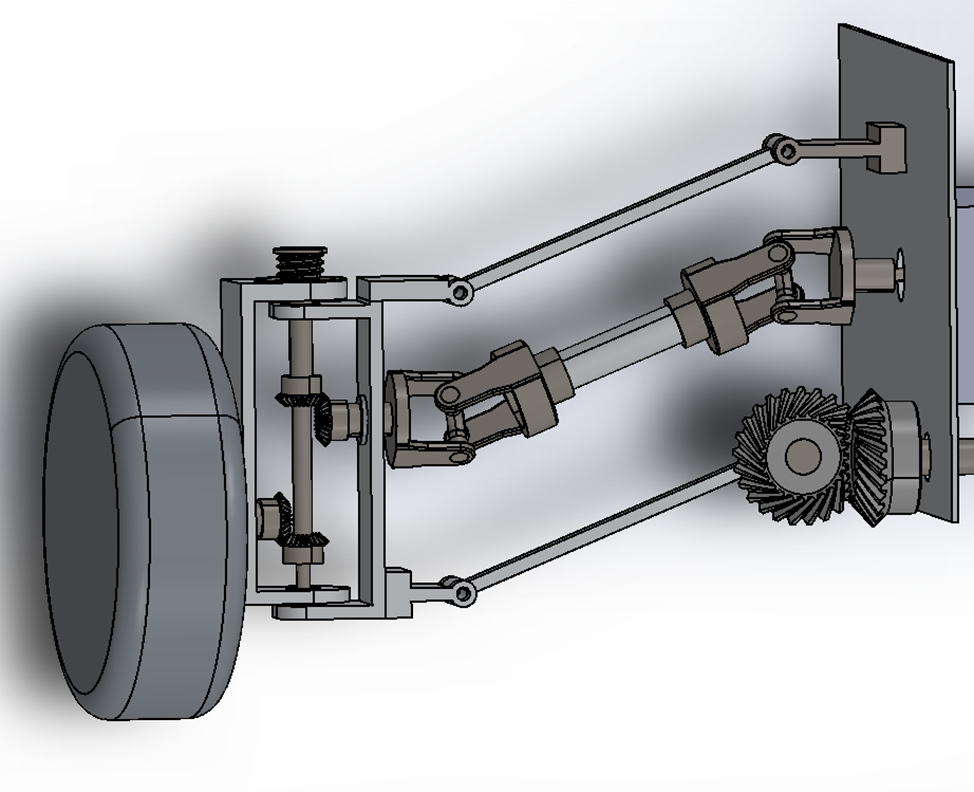}
    \caption{DISTANT Flight Model design}
    \label{fig:distant_fm_design}
\end{figure}

Planetary exploration missions continue to push the boundaries of autonomous surface operations, with rovers increasingly required to traverse greater distances and operate for extended periods in harsh environments \cite{inotsume_parametric_2019, barthelmes_mmx_2022}. Current rover designs typically employ hub-mounted motors or chassis-embedded drive systems that expose critical actuators to extreme thermal cycling, dust infiltration, and mechanical stress during long-traverse missions \cite{von_ehrenfried_perseverances_2022, ding_2-year_2022}.

The European Space Agency has identified the need for advanced locomotion systems capable of supporting missions requiring traversals of 50 km or more without significant performance degradation. Traditional approaches face limitations in dust protection, thermal management, and component longevity that become critical factors in extended operations \cite{shaw_mineral_2022}.

The DISTANT project addresses these challenges through a paradigm shift in rover architecture, relocating all traction and steering actuators to a thermally controlled warm box within the rover body. This approach enables centralised thermal management, simplified dust protection, and enhanced maintenance accessibility whilst maintaining full locomotion control capabilities. The DISTANT design (Figure \ref{fig:distant_fm_design}) employs multiple transmission mechanisms including cardan joints, bevel gears, and capstan drives to control suspension, steering, and wheel rotation from the protected warm-box environment.

\section{STATE OF THE ART}

\subsection{Current Rover Locomotion Systems}

Contemporary planetary rovers employ various locomotion architectures, each with distinct advantages and limitations for extended missions. The most prevalent configurations include rocker-bogie suspension systems, as demonstrated by the Mars Exploration Rovers and Perseverance \cite{von_ehrenfried_perseverances_2022}, and independent suspension designs utilised in advanced concepts such as the European Moon Rover System (EMRS) \cite{luna_european_2023}.

Rocker-bogie systems provide exceptional obstacle traversal capabilities, enabling rovers to navigate obstacles up to twice their wheel diameter whilst maintaining ground contact across all wheels \cite{d_s_design_2017}. This passive suspension design has proven reliability for planetary surface operations, though it limits active suspension control and centre of mass adjustment capabilities.

Independent suspension systems, particularly double wishbone configurations, offer superior terrain adaptability through active control of individual wheel positions \cite{luna_breadboarding_2024}. These systems enable dynamic centre of mass adjustment, improving stability on slopes and allowing scientific instrument positioning without additional robotic mechanisms.

\subsection{Transmission System Technologies}

Current transmission technologies for rover applications include hub-mounted motors, chassis-embedded drive systems, and various mechanical transmission mechanisms. Hub-mounted configurations provide direct wheel control but expose motors to environmental extremes \cite{stockman_efficiency_2015}. Chassis-embedded systems offer partial protection whilst requiring complex transmission routes to wheels.

Mechanical transmission options include cardan joints for angular flexibility, belt drive systems for distributed power transmission, and gear-based solutions for high torque applications \cite{sherr-thoss_universal_1998}. Each technology presents trade-offs between efficiency, complexity, environmental resistance, and maintenance requirements.

\subsection{Environmental Protection Challenges}

Dust contamination represents a primary concern for mechanical systems on planetary surfaces, particularly in lunar environments where electrostatic charging exacerbates particle adhesion \cite{dong_parameters_2017, Engelschiøn2020, joulaud_investigation_2024}. Traditional approaches to dust mitigation include sealed enclosures, protective covers, and air filtration systems, though these solutions add mass and complexity whilst providing limited effectiveness for moving parts.

Thermal cycling poses additional challenges, with temperature variations exceeding 300 K on lunar surfaces causing material expansion, seal degradation, and lubrication breakdown \cite{reitz_radiation_2012}. Current thermal protection methods include Multi-Layer Insulation (MLI), Surface Layer Insulation (SLI), and active heating systems, each with specific applications and limitations.

\section{SYSTEM REQUIREMENTS}

The DISTANT system addresses mission-critical requirements for extended planetary surface operations through comprehensive functional and performance specifications.

\subsection{Mission Requirements}

The primary mission requirement mandates 50 km traverse capability without performance degradation, necessitating robust mechanical design and environmental protection strategies. The system must maintain full locomotion control including wheel traction, steering, speed regulation, and position feedback through integrated sensor systems.

Operational requirements include compatibility with multiple locomotion modes: point turning, crab steering, and Ackermann steering configurations. The design must accommodate steering angles of ±90 degrees minimum, with capability for full 360-degree rotation where architecturally feasible.

\subsection{Physical Constraints}

Dimensional constraints limit assembled wheel radius to 35 cm maximum, derived from launch vehicle accommodation requirements and scaling considerations from EMRS heritage designs. The wheel width specification of 11 cm reflects optimisation for traction performance whilst maintaining reasonable ground pressure on regolith surfaces.

Mass distribution requirements favour centralised actuator placement to improve rover stability and enable dynamic centre of mass adjustment. This approach reduces unsprung mass at wheel locations, improving suspension responsiveness and reducing impact loads during traverse operations.

\subsection{Environmental Specifications}

Thermal protection requirements address operational temperature ranges from -180°C to +120°C, covering lunar polar and equatorial conditions. The warm box environment must maintain actuator temperatures within operational limits through active thermal control systems.

Dust protection specifications target particle exclusion from all moving mechanical interfaces, with particular emphasis on bearing surfaces, transmission joints, and actuator mechanisms. The design incorporates protective bellows, sealed enclosures, and filtered ventilation where required.

\section{DESIGN APPROACH}

\subsection{Architectural Philosophy}

The DISTANT design philosophy prioritises component protection through centralisation whilst maintaining full locomotion capabilities. All primary actuators are housed within a thermally controlled warm box, connected to wheels through protected transmission systems. This approach enables simplified thermal management, coordinated dust protection, and enhanced maintenance accessibility.

The modular design strategy facilitates component replacement, system scaling, and mission-specific adaptations. Independent suspension legs in double wishbone configurations enable incremental testing and validation, whilst maintaining system-level coordination benefits.

\subsection{Transmission Strategy}

Transmission system selection balances mechanical efficiency, environmental protection, and system complexity considerations. Cardan joint systems provide angular flexibility for suspension articulation whilst maintaining power transmission continuity. Capstan drive configurations enable distributed power routing with inherent overload protection through controlled slip mechanisms.

The integrated approach combines multiple transmission technologies to optimise performance across different operational scenarios. Wheel traction utilises cardan joints for suspension accommodation, whilst steering transmission employs capstan drive for precise control with mechanical advantage adjustment through drum diameter selection.

\section{DESIGN TRADE-OFFS}

The four proposed configurations represent distinct design philosophies, each balancing mechanical complexity, terrain adaptability, efficiency, and reliability differently.

\textbf{Model 1} combines a \textbf{double wishbone suspension with capstan steering}, offering high kinematic and mechanical efficiency in a lightweight and modular package. Its reduced mass and compact dimensions make it ideal for versatile locomotion and integration. However, cable tensioning in the capstan system introduces wear and maintenance challenges.

\textbf{Model 2} shares the same suspension and wheel transmission as Model 1 but replaces the capstan with a \textbf{gearbox-based steering system}. This enhances mechanical reliability and avoids cable degradation, though at the cost of increased weight, reduced steering articulation, and potential alignment issues.

\textbf{Model 3} employs a \textbf{rocker-bogie suspension with belt-driven transmission}, excelling in terrain adaptability and passive shock absorption. While its mobility is high, it is significantly heavier and more mechanically complex. Its reliance on belts and pulleys increases the number of failure points and complicates maintenance.

\textbf{Model 4} improves upon Model 3 by introducing a \textbf{hybrid gear-belt transmission}, reducing belt-related failures and improving reliability. Nonetheless, it retains the high weight and complexity of the rocker-bogie structure, with limited modularity compared to Models 1 and 2.

In summary, Models 1 and 2 prioritize simplicity, modularity, and weight efficiency, making them more suitable when payload capacity and power efficiency are critical. Models 3 and 4 favour terrain adaptability and passive compliance but introduce mass and complexity that can compromise performance in constrained missions. Among all, Model 1 demonstrates the most balanced trade-off, combining low weight, high efficiency, and adaptability with acceptable mechanical risks.

\section{SELECTED DESIGN CONFIGURATION}

\begin{figure}[h!]
\centering
\includegraphics[width=1\linewidth]{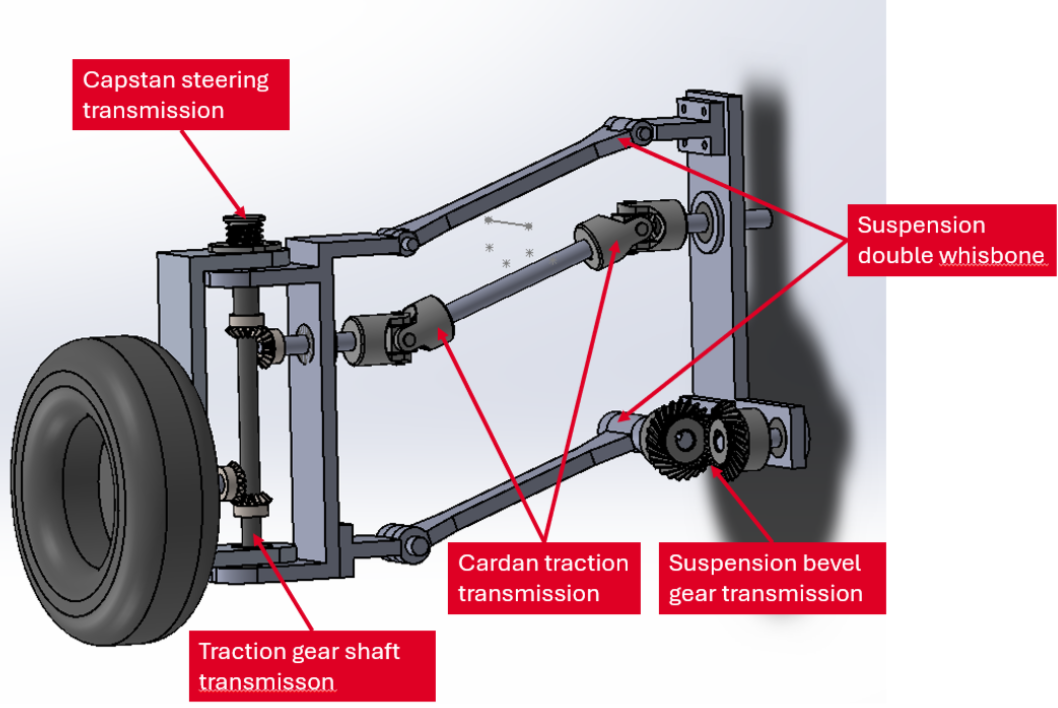}
\caption{Selected DISTANT configuration showing double wishbone suspension with cardan joint wheel transmission and capstan drive steering system. \label{fig:model1}}
\label{fig:distant_architecture}
\end{figure}

Following comprehensive analysis of four alternative architectures, the double wishbone suspension configuration with cardan joint wheel transmission and capstan drive steering control (Model 1) was selected as the optimal solution (Figure \ref{fig:distant_architecture}).

\subsection{Architecture Overview}

The selected architecture employs a double wishbone suspension configuration with cardan joint wheel transmission and capstan drive steering control. The design utilises a parallelogram suspension structure enabling independent wheel height adjustment for terrain adaptation and centre of mass control \cite{luna_breadboarding_2024}.

Wheel traction transmission incorporates dual cardan joints arranged in series to accommodate suspension articulation and steering rotation with minimal efficiency loss. The parallelogram geometry ensures equal but opposite angular deflections across cardan pairs, optimising transmission efficiency through balanced loading.

Steering control utilises a capstan drive system with steel cable transmission from the warm box to wheel assemblies. This configuration provides precise angular control with mechanical advantage adjustment through drum diameter selection. Cable routing follows protected paths within suspension structure members, minimising environmental exposure.

\subsection{Performance Characteristics}

The selected design demonstrates superior performance characteristics compared to alternative configurations. Wheel transmission efficiency ranges from 43-99\% depending on articulation angles, with steering efficiency reaching 91\%. Torque requirements span 30.01-86.11 Nm for wheel drive and 26.01-32.09 Nm for steering operations under various terrain conditions.

The total architecture mass of 10.25 kg provides favourable mass distribution with maximum range of 738 mm from warm box to wheel connection. The design accommodates suspension range of -5° to +30° with steering angle capability of ±90°, meeting mission mobility requirements.

\subsection{Design Rationale}

Model 1 was selected based on comprehensive trade-off analysis considering mechanical performance, environmental protection, and operational flexibility. The configuration demonstrates optimal efficiency in wheel transmission due to balanced cardan joint arrangements and reduced angular deflections compared to alternative designs.

The capstan steering system provides higher efficiency compared to gearbox alternatives whilst maintaining precise control capabilities. The design minimises weight and complexity whilst offering superior modularity for component replacement and system scaling. The double wishbone configuration enables both passive and active suspension modes with excellent terrain adaptability and centre of mass adjustment capabilities.

\section{DESIGN ANALYSIS}

\subsection{Mechanical Performance}

Torque analysis indicates optimal performance characteristics for the selected configuration. Wheel drive requirements of 30.01-86.11 Nm reflect terrain-dependent loading conditions, whilst steering torque requirements of 26.01-32.09 Nm provide adequate control authority across operational scenarios.

The double wishbone configuration demonstrates superior efficiency in wheel transmission due to optimised cardan joint arrangements and reduced angular deflections. Wheel transmission efficiency of 43-99\% depends on articulation angles, with steering efficiency reaching 91\% under nominal conditions.

The parallelogram suspension geometry ensures balanced loading across cardan joint pairs, maximising transmission efficiency whilst accommodating full suspension articulation range. This approach minimises mechanical stress and wear compared to alternative transmission configurations.

\subsection{Environmental Protection}

Dust protection strategies utilise protective bellows at cardan joints and sealed cable routing within suspension members. The modular architecture enables localised protection systems tailored to specific component requirements.

Advanced coating technologies including lotus leaf coatings \cite{calle_active_2011} and electrostatic discharge solutions offer supplementary protection mechanisms. These approaches complement mechanical barriers through surface treatment strategies that reduce particle adhesion and enable self-cleaning capabilities.

Thermal analysis indicates that centralised actuator placement enables effective thermal management through the warm box environment. Temperature modelling shows maintained actuator temperatures within operational ranges across lunar thermal cycles with appropriate insulation and active heating systems.

\subsection{Reliability and Maintainability}

Reliability assessment identifies low failure risk due to reduced component count and simplified transmission paths. Cardan joint bearing wear and cable degradation represent primary failure mechanisms requiring design attention through material selection and maintenance protocols.

Modularity analysis favours the double wishbone configuration for its independent leg design enabling incremental testing, component replacement, and system scaling. The architecture provides operational flexibility whilst maintaining maintenance accessibility through modular construction.

Critical failure modes include bearing wear in cardan joints, cable degradation in capstan systems, and efficiency loss at complex joint positions. Design margins and redundancy strategies address these concerns through material selection and operational monitoring systems.

\section{BREADBOARD DESIGN}

The selected Model 1 configuration has been refined and scaled to 1:3 ratio according to test facility constraints and scaling analysis requirements.

\begin{figure}[h!]
    \centering
    \includegraphics[width=1\linewidth]{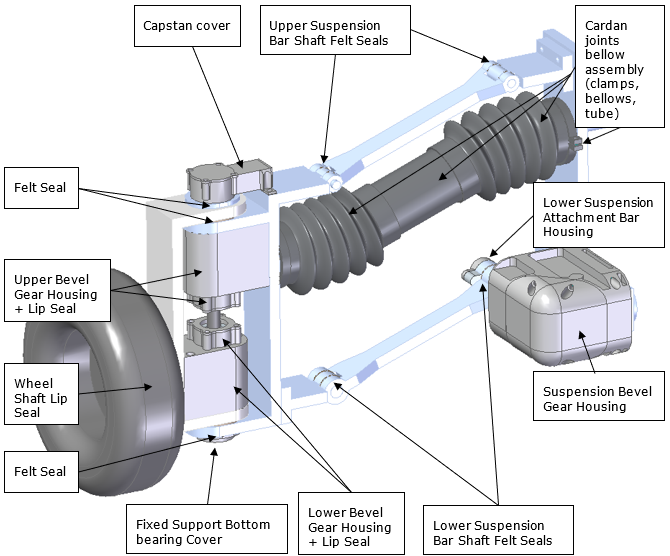}
    \caption{Dust isolation CAD design showing protective bellows, seals, and contamination barriers}
    \label{fig:distant_dust}
\end{figure}

The breadboard configuration incorporates comprehensive dust isolation mechanisms as shown in Figure~\ref{fig:distant_dust}. The scaled design maintains critical functionality whilst accommodating testing facility dimensional constraints and enabling detailed component validation.

\subsection{Breadboard Component Description}

\subsubsection{Structural and Transmission System}

The breadboard incorporates scaled versions of all critical transmission components. Cardan joints provide power transmission with accommodation for suspension and steering articulation. The dual cardan configuration ensures balanced angular loading whilst maintaining transmission efficiency across operational ranges.

The warm box houses all motorisation systems, providing centralised thermal control and dust protection. Actuator integration includes wheel drive motors, steering control systems, and suspension positioning mechanisms. Interface connections utilise protected routing through sealed penetrations in the warm box structure.

Material selection emphasises space-qualified options including aluminium 7075-T6 for structural components, stainless steel 304 for bellows assemblies, and PTFE for sealing applications. Component specifications include steel AISI 3115 for high-stress mechanical elements and titanium for specialised applications requiring low thermal expansion.

\subsubsection{Dust Isolation Components}

The dust protection system incorporates multiple barrier types adapted for breadboard scale constraints. Felt seals provide contamination barriers for translational movement interfaces, whilst lip seals protect rotational components. Protective bellows cover cardan joint assemblies, preventing particle ingress during articulation.

Critical dust isolation components include suspension bar felt seals, bearing covers, bevel gear housings with integrated seals, and comprehensive capstan protection systems. The design maintains flight model protection principles whilst accommodating manufacturing constraints imposed by breadboard scaling.

Component specifications include:
\begin{itemize}
\item 8 suspension bar felt seals for translational movement protection
\item 2 bevel gear housings with integrated lip seals (6x16 PTFE)
\item 2 cardan protective bellows with stainless steel construction
\item 1 capstan assembly with dedicated felt seal protection
\item Multiple bearing covers and attachment housings
\end{itemize}

\subsubsection{Thermal Components}

The thermal control system addresses breadboard testing requirements within thermal vacuum chamber constraints. The warm box incorporates Multi-Layer Insulation (MLI) for radiative heat exchange control, supplemented by active heating elements with integrated temperature sensors.

Thermal design considerations address Dirty Thermal Vacuum Chamber (DTVC) limitations, particularly the absence of internal cooling loops which constrains hot case testing scenarios. MLI configuration reduces thermal exchange with both environmental conditions and the breadboard mechanical systems, enabling representative thermal control validation.

Heater integration provides temperature regulation during both hot and cold thermal cycles. The system maintains warm box temperatures within operational ranges whilst accommodating chamber thermal setpoint constraints during testing campaigns.

\subsection{Warm-Box design}

\begin{figure}[h!]
    \centering
    \includegraphics[width=1\linewidth]{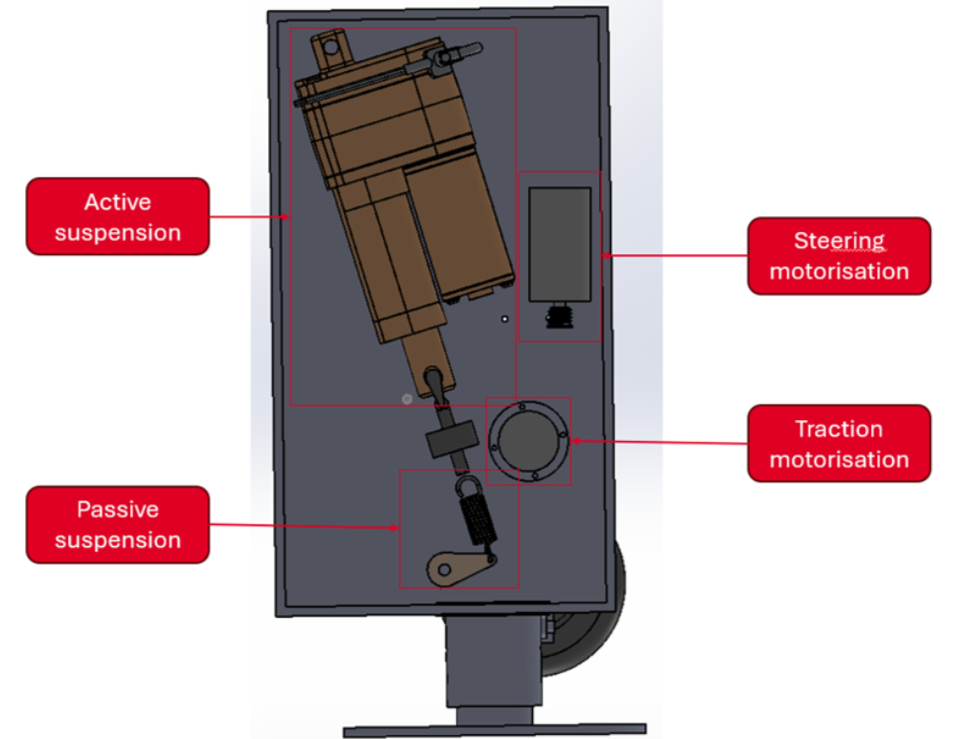}
    \caption{Warm-Box design including active suspension and passive suspension systems, steering motorisation and traction motorisation}
    \label{fig:warmbox}
\end{figure}

The warm-box design (Figure \ref{fig:warmbox}) features a hybrid suspension system that combines passive and active elements for terrain adaptation and load distribution. The system includes a spring for passive suspension and a linear actuator for active control. The design also houses the traction and steering motorisation systems within the thermally controlled environment.

The active suspension transmission is located at the bottom of the parallelogram structure, with the rotation point positioned along a horizontal axis near the warm box. This axis is driven by a bevel gear system controlled by an actuator within the chassis. The rotational motion controls the angle of the connected bar relative to the ground, with both steering and traction systems attached via a mounting plate at the bar's end. This parallelogram mechanism allows the wheel's end connection to move vertically, enabling the suspension to absorb terrain vibrations and shock loads. The traction system uses dual cardan joints to transmit motion from the actuator to the wheel, whilst the steering system employs a capstan drive mechanism with tensioned cables that adjust the steering angle through drum rotation.

\section{MATERIALS AND MANUFACTURING}

\subsection{Material Selection}

Material selection addresses the unique requirements of planetary surface operations including temperature extremes, radiation exposure, and mechanical stress cycling. Primary structural materials include aluminium 7075-T6 for lightweight structural components, steel AISI 3115 for high-stress mechanical elements, and titanium for specialised applications requiring low thermal expansion and high strength-to-weight ratios.

Steel selection for gears, cardan joints, and cable components reflects requirements for high tensile strength, fatigue resistance, and dimensional stability under thermal cycling. Aluminium allocation to structural members and pulley assemblies optimises mass distribution whilst providing adequate strength for operational loads.

Advanced materials investigation includes bulk metallic glasses for exceptional mechanical properties and wear resistance \cite{browne_metallic_2022}. These materials offer potential performance improvements for bearing applications and high-stress transmission components, though manufacturing complexity requires careful evaluation.

\subsection{Manufacturing Considerations}

Manufacturing approaches emphasise proven space-qualified processes and materials compatibility with existing rover fabrication capabilities. Mechanical components utilise conventional machining, welding, and assembly techniques adapted for space applications.

Quality control procedures address dimensional tolerances critical for transmission efficiency and environmental sealing effectiveness. Particular attention focuses on cardan joint clearances, gear tooth profiles, and cable routing geometry to ensure reliable operation across temperature ranges.

Assembly procedures accommodate modular testing and validation requirements whilst maintaining contamination control throughout manufacturing and integration phases.

\section{VALIDATION APPROACH}

\subsection{Testing Strategy}

The validation approach follows a structured progression from component-level verification through integrated system testing. Manufacturing, Assembly, Integration, and Testing (MAIT) activities commence following Breadboard Design Review approval, establishing the foundation for comprehensive system validation.

Component-level testing addresses individual subsystem performance including actuator characterisation, transmission efficiency measurement, and environmental protection validation. Subsystem integration testing verifies interface compatibility and coordinated operation before full system assembly.

System-level validation encompasses functional performance verification, environmental testing, and operational scenario demonstration. The testing sequence ensures incremental validation whilst minimising integration risks through controlled progression.

\subsection{Test Facilities and Equipment}

Validation activities require specialised facilities including cleanroom environments for contamination-sensitive integration, thermal vacuum chambers for environmental testing, and regolith simulation capabilities for mobility assessment.

Spacive's thermal vacuum chambers provide primary environmental testing capabilities. The clean thermal vacuum chamber achieves $9.5×10^{-7}$ mbar pressure with temperature control from -180°C to +150°C. The dirty vacuum chamber accommodates regolith testing with $9.5×10^{-3}$ mbar pressure capability and 500 mm diameter testing volume.

Ground support equipment development addresses unique requirements for DISTANT system testing including custom fixtures for transmission alignment, specialised instrumentation for torque and efficiency measurement, and environmental simulation systems for dust and thermal testing.

\subsection{Performance Metrics}

Success criteria encompass functional performance parameters, environmental tolerance limits, and operational reliability measures. Primary metrics include transmission efficiency maintenance across temperature ranges, dust exclusion effectiveness, and mechanical wear rates during extended operation testing.

Quantitative performance targets include 95\% transmission efficiency maintenance, contamination exclusion to specified particle size limits, and operation through 50 km equivalent traverse cycles without performance degradation. These metrics provide objective validation criteria for design acceptance.

Operational testing scenarios address realistic mission profiles including obstacle traversal, slope navigation, and extended traverse operations. Performance assessment across these scenarios validates design adequacy for intended mission applications.

\section{IMPLEMENTATION TIMELINE}

The DISTANT project follows a phased development approach with clearly defined milestones and deliverables. Following completion of the preliminary design phase in mid-2025, activities transitioned to detailed design finalisation and breadboard design.

Breadboard manufacturing and assembly activities are scheduled for Q4 2025, incorporating lessons learned from preliminary design analysis and trade-off studies. Manufacturing emphasises space-qualified processes and materials whilst maintaining cost-effectiveness for demonstration purposes.

Integration and testing activities commence in Q1 2026 following Test Readiness Review completion. The testing campaign addresses both component-level validation and integrated system performance assessment across relevant environmental conditions and operational scenarios.

System validation activities extend through Q2 2026, culminating in design maturity assessment and recommendations for flight system development. The timeline accommodates iterative design refinement based on testing results whilst maintaining overall programme schedule objectives.

\section{FUTURE WORK}

\subsection{Breadboard Validation}

The immediate focus centres on breadboard manufacturing completion and comprehensive testing validation. The Q1 2026 testing campaign will provide critical performance data for transmission efficiency, environmental protection effectiveness, and mechanical reliability under simulated operational conditions.

Testing results will inform any necessary design refinements whilst demonstrating technology readiness advancement. Specific validation objectives include dust exclusion performance assessment, thermal management effectiveness evaluation, and long-duration operation reliability demonstration.

Component-level testing addresses individual subsystem performance including actuator integration, transmission efficiency measurement across operational ranges, and environmental protection validation under dust and thermal conditions. The modular breadboard architecture enables isolated component testing whilst maintaining system-level integration verification.

\subsection{Flight System Development}

Flight system development activities await breadboard validation completion and design maturity assessment. The progression from breadboard demonstration to flight-qualified hardware requires systematic validation of materials, processes, and performance characteristics under mission-relevant conditions.

Advanced materials investigation addresses next-generation solutions for improved environmental resistance, mass reduction, and enhanced mechanical properties. Bulk metallic glasses, advanced ceramics, and composite materials offer potential performance improvements for specialised applications requiring extreme environmental tolerance.

Manufacturing process optimisation focuses on cost reduction, quality improvement, and production scalability for multiple mission applications. The transition from development hardware to operational systems requires process automation advancement and quality control enhancement to support mission implementation schedules.

\subsection{Technology Maturation}

Technology maturation activities address component qualification completion, environmental testing extension, and operational validation across comprehensive mission scenarios. The progression from technology demonstration to mission-ready systems requires systematic validation of reliability, durability, and performance consistency.

System integration optimisation targets design simplification, reliability improvement, and maintainability enhancement through design evolution informed by testing experience. Operational procedures development addresses deployment protocols, maintenance requirements, and troubleshooting methodologies for mission implementation.

The DISTANT technology foundation enables application across multiple mission contexts including lunar surface exploration, Mars operations, and asteroid proximity missions. System adaptability supports mission-specific requirements whilst maintaining core technology benefits of centralised actuator protection and simplified thermal management.

\section*{ACKNOWLEDGMENTS}

The DISTANT project is developed with the support of the European Space Agency (ESA), funding the DISTANT project development under contract No. 4000146625/24/NL/RK.

\addcontentsline{toc}{chapter}{References}
\small
\printbibliography

\end{document}